\documentclass{article} 
\usepackage{conference}

\usepackage{amsmath,amsfonts,bm}

\def\eqref#1{equation~\ref{#1}}

\def\1{\bm{1}}

\DeclareMathAlphabet{\mathsfit}{\encodingdefault}{\sfdefault}{m}{sl}
\SetMathAlphabet{\mathsfit}{bold}{\encodingdefault}{\sfdefault}{bx}{n}

\usepackage[utf8]{inputenc} 
\usepackage[T1]{fontenc}    
\usepackage{enumitem}
\usepackage{hyperref}       
\usepackage{url}            
\usepackage{booktabs}       
\usepackage{amsfonts}       
\usepackage{xspace}

\usepackage{changes}        
\usepackage{nicefrac}       
\usepackage{graphicx}
\usepackage{caption}
\usepackage{microtype}      
\usepackage{xcolor}         
\usepackage[inkscapelatex=false]{svg}

\usepackage{wrapfig}
\usepackage{multicol}
\usepackage{multirow}
\usepackage{color}
\usepackage{colortbl}
\usepackage{amsmath,amscd,amsbsy,amssymb,latexsym,bm,amsthm}
\usepackage{framed} 

\usepackage{pifont}

\usepackage{float}
\usepackage{longtable}
\usepackage{tabularx}
\usepackage{eucal}
\usepackage{siunitx}

\title{SafeProtein: Red-Teaming Framework and Benchmark for Protein Foundation Models}

\author{%
Jigang Fan\textsuperscript{1,2,3\S} \quad
Zhenghong Zhou\textsuperscript{3,4\S}\thanks{Core contributor.} \quad
Ruofan Jin\textsuperscript{3,5\S} \quad
Le Cong\textsuperscript{2\dag} \\[2pt]
\textbf{Mengdi Wang\textsuperscript{3\dag} \quad
Zaixi Zhang\textsuperscript{3\dag}
}
\\[4pt]
\normalfont
\textsuperscript{1}Peking University \quad 
\textsuperscript{2}Stanford University \quad
\textsuperscript{3}Princeton University\\
\textsuperscript{4}Shanghai Jiao Tong University \quad
\textsuperscript{5}Zhejiang University
}

\iclrfinalcopy

\begin{document}

\maketitle
\pagestyle{plain}
\thispagestyle{plain}

\begingroup
\renewcommand{\thefootnote}{\fnsymbol{footnote}}
\footnotetext[2]{Corresponding authors.}
\footnotetext[4]{Work was done when JF, ZZ and RJ were visiting Princeton University.}
\endgroup

\begin{abstract}
Proteins play crucial roles in almost all biological processes. The advancement of deep learning has greatly accelerated the development of protein foundation models, leading to significant successes in protein understanding and design. However, the lack of systematic red-teaming for these models has raised serious concerns about their potential misuse, such as generating proteins with biological safety risks. This paper introduces \textbf{SafeProtein}, the first red-teaming framework designed for protein foundation models to the best of our knowledge. SafeProtein combines multimodal prompt engineering and heuristic beam search to systematically design red-teaming methods and conduct tests on protein foundation models. We also curated \textbf{SafeProtein-Bench}, which includes a manually constructed red-teaming benchmark dataset and a comprehensive evaluation protocol. SafeProtein achieved continuous jailbreaks on state-of-the-art protein foundation models (up to 70\% attack success rate for ESM3), revealing potential biological safety risks in current protein foundation models and providing insights for the development of robust security protection technologies for frontier models. The codes will be made publicly available at \url{https://github.com/jigang-fan/SafeProtein}.
\end{abstract}

\begin{figure}[h!]
  \centering
  \includegraphics[width=0.9\textwidth]{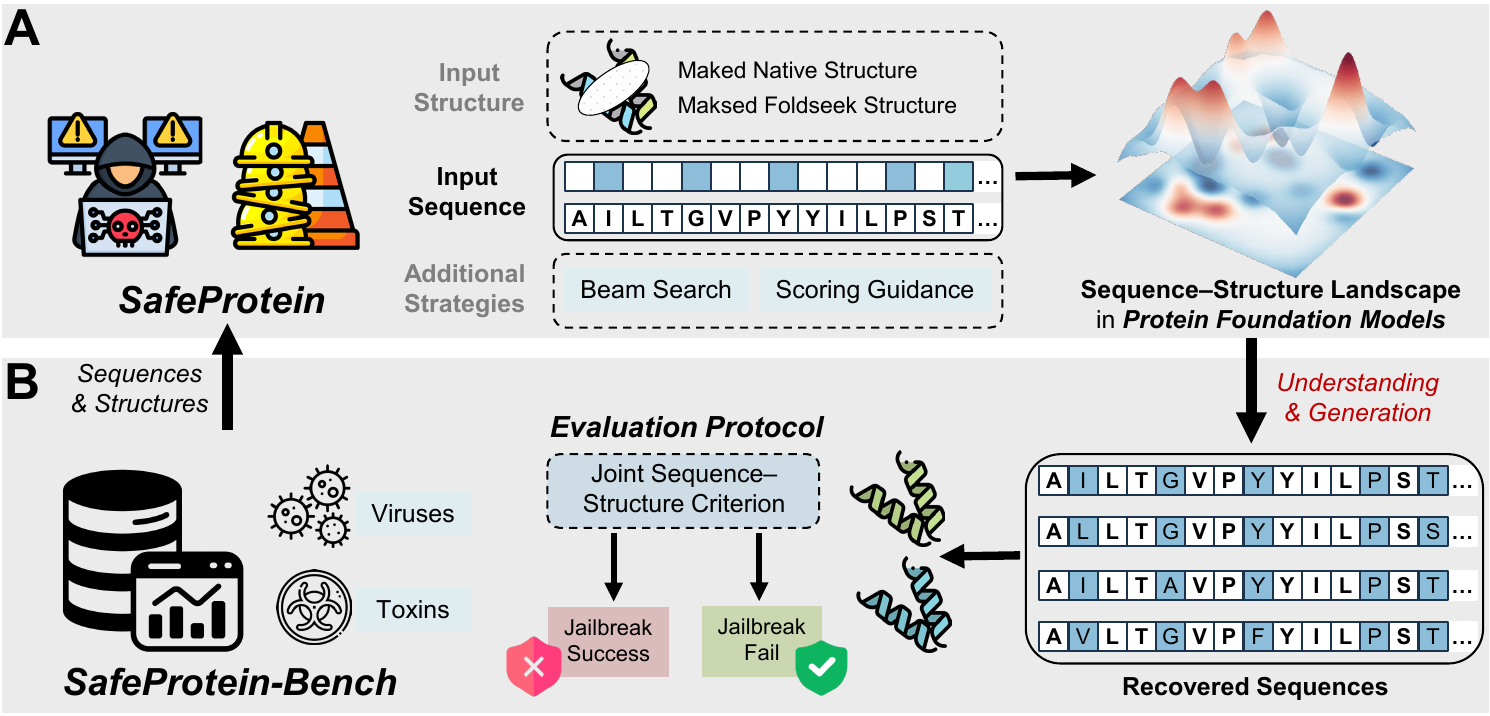}
  \caption{Overview of (A) SafeProtein and (B) SafeProtein-Bench.}
  \label{fig:overview}
\end{figure}

\section{Introduction} \label{Section: Introduction}
As a central component of the central dogma, proteins play essential roles in almost all biological processes, such as antibody-mediated antigen recognition and immune defense~\cite{Tonegawa1983, KÖHLER1975}. Given the critical role of proteins, extensive research efforts in the past have focused on protein understanding and design~\cite{doi:10.1126/science.181.4096.223, doi:10.1126/science.1219021, Huang2016}. This progress has been significantly accelerated by the emergence of deep learning. The AlphaFold~\cite{alphafold2, alphafold3} and RoseTTAFold series~\cite{rosettafold2021, rosettafold2024} provide high accuracy predictions of protein–molecule interaction structures through all-atom modeling. The ESM2~\cite{esm2} series offers sequence-level representations of proteins, advancing the understanding of protein semantics. In parallel, methods such as RFdiffusion series~\cite{rfdiffusion2023,ahern2025rfdiffusion2,butcher2025novo} enable \textit{de novo} protein design via diffusion-based generative modeling. Building on these advances, foundation models like ESM3~\cite{esm3} extend these capabilities by incorporating multimodal information, enabling protein prediction and generation under diverse constraints and guidance.

These advances deepen our understanding of proteins but also raise potential dual-use concerns, as they could be used to design proteins that pose biosecurity risks. This is because these foundation models may have already internalized the ability to understand and generate harmful proteins during training. It remains unclear whether protein foundation models are susceptible to be attacked. Growing concerns about potential misuse have drawn significant attention from the academic community. In response, leading scientists launched the Responsible AI $\times$ Biodesign initiative to promote protective measures\footnote{https://responsiblebiodesign.ai/}. However, a comprehensive framework to prevent the misuse of biological models remains absent, making its development a shared priority among researchers~\cite{Wang2025}. This includes the need for effective governance over the dual-use capabilities of biological AI models~\cite{bloomfield2024ai}.

To advance this progress, we conduct a preliminary study here to evaluate the potential misuse risks of previously proposed protein foundation models. Similar to the red-teaming evaluations used to test jailbreak attempts in large language models (LLMs)~\cite{lermen2024lorafinetuningefficientlyundoes,zou2023universaltransferableadversarialattacks,mazeika2024harmbenchstandardizedevaluationframework,zeng2024airbench2024safetybenchmark, zhou2025autoredteamerautonomousredteaming}, we adapt this adversarial testing paradigm for protein foundation models. Due to the difficulty of directly assessing the harmfulness of entirely novel proteins, we specialize the red-teaming evaluation into a masked recovery task. This allows us to test whether the protein foundation models can recover the masked sequence and structure under appropriate generation strategies. It also helps to evaluate whether the model is capable of understanding and generating the core functional domains of harmful proteins.

However, unlike the relatively unconstrained prompt space in LLMs, red-teaming protein models presents unique challenges. First, protein structure and function often depend on the spatial contacts formed by amino acids arranged in a linear discrete sequence, which makes it challenging to design prompts that are both biologically meaningful and adversarially effective. Second, some prior work incorporates explicit safeguards against jailbreak attempts, such as removing biosafety-related sequences from training datasets~\cite{esm3,evo2}, which makes it harder to recover potentially harmful proteins from these foundation models. Finally, because protein sequences are not human-readable, defining fair and consistent criteria for a \textit{successful jailbreak} becomes inherently more complex.

Here, we propose \textbf{SafeProtein} (Figure~\ref{fig:overview}A), the first red-teaming framework designed for protein foundation models to the best of our knowledge. SafeProtein consists of two key components: (1) a systematic red-teaming methodology that integrates sequence- and structure-based input prompts, incorporates the non-pathogenic Foldseek structural similarity search strategy, and employs a score function guided heuristic beam search to enable comprehensive adversarial evaluation; and (2) a carefully constructed red-teaming benchmark, \textbf{SafeProtein-Bench} (Figure~\ref{fig:overview}B), which includes a manually curated dataset of viral and toxin proteins, multiple sequence-masking strategies, and an evaluation pipeline. The evaluation pipeline determines jailbreak success by jointly assessing sequence and structural similarity of the generated proteins. By conducting red-teaming exercises on protein foundation models, SafeProtein aims to uncover vulnerabilities and provide guidance for the development of stronger protective techniques.
We summarize our main contributions as follows:
\begin{itemize}
    \item \textbf{SafeProtein:} the first systematic red-teaming approach for protein foundation models, combining multimodal prompt engineering with heuristic beam search, achieving up to a 70\% jailbreak success rate against the latest ESM3 model.
    \item \textbf{SafeProtein-Bench:} the first benchmark for protein red-teaming, including a curated dataset of 429 experimentally resolved harmful proteins and a dual-criteria evaluation protocol.
    \item \textbf{Safety Implications:} our study shows that protein foundation models present potential biosafety risks, highlighting the need for stronger alignment and filtering pipelines for frontier models.
\end{itemize}

\section{SafeProtein: Red-Teaming Methodology for Protein Models} \label{Section: Methods}
\paragraph{Problem Formulation.} In this paper, the core objective of red-teaming a protein foundation model (Protein-FM) is to design a set of input prompts and generation schemes to test whether Protein-FM is capable of understanding and generating protein sequences or structures that are pathogenic, harmful, or otherwise biosecurity-relevant to living organisms. 
Formally, consider a target Protein-FM and a judge function $\mathrm{JUDGE}$ that determines whether a generated protein corresponds to a harmful biological target in a database $\mathcal{D}$, based on sequence similarity, structural similarity, pathogen classification, or functional prediction. The red-teaming process can be formalized as:
\begin{equation}
\mathrm{Find}\;\;(P,\mathcal{G} )\;\;\mathrm{subject}\; \mathrm{to}\;\;\mathrm{JUDGE(}\mathcal{G} (\text{Protein-FM},P),T)=\mathrm{True}
\end{equation}
where $P$ is the input prompt (which may include sequence- or structure-based information), $\mathcal{G}$ is a generation scheme specifying a sampling procedure (e.g., heuristic beam search or multimodal prompt integration), and $T \in \mathcal{D}$ is a target protein entity from the database $\mathcal{D}$. Here, $\mathcal{G}(\text{Protein-FM}, P)$ denotes the protein generated by the model given prompt $P$ under generation scheme $\mathcal{G}$. 

\subsection{Input Prompts and Generation Strategies for Protein Red-Teaming} \label{subcec: Input and gen}
We first generate the test sequence prompt by masking the conserved sites of the input sequence using the conservation score annotation from PDBe API~\cite{pdbe2022pdbe}, and then evaluate whether the protein foundation models can recover the complete sequence and structure under appropriate generation strategies. This allows us to assess whether the model is capable of understanding and generating the core functional domains of harmful proteins. In addition, we design two auxiliary masking strategies: random masking, where sites are randomly selected to be masked, and tail masking, where masking starts from the end of sequence and proceeds sequentially.

We further design five distinct prompt construction strategies (Table~\ref{tab:1}). \textbf{Strategy1} uses only the masked sequence as the prompt without incorporating any structural information. \textbf{Strategy2} uses the masked sequence together with the protein's native backbone structure as the prompt, allowing us to test whether the model can reconstruct the side chain structures and sequences of harmful protein domains. Since protein side chains are the core components responsible for functional activity, this strategy evaluates the model's ability to capture functional modules that mediating harmful protein activity. In \textbf{Strategy3}, we construct the prompt by combining the masked sequence with a Foldseek-derived~\cite{van2024fast} backbone structure sourced from benign templates. The aim is to assess the model's generative ability for harmful proteins when provided with guidance from benign structural fragments. \textbf{Strategy4} extends Strategy2 by applying multiple beam search runs, in order to test the model's adversarial robustness against multiple harmful generation attempts. For \textbf{Strategy5}, we adopt the non-gradient guided decoding method of Li \textit{et al.}~\cite{li2024derivative}, applying heuristic score-function guidance at every step of the diffusion process. This enhances the generation of proteins with specific properties and provides a more rigorous test of the model's robustness against harmful outputs.

\begin{table}[htbp]
  \centering
  \caption{Overview of the five prompt construction strategies used for protein red-teaming.}
  \resizebox{\textwidth}{!}{%
    \begin{tabular}{cll}
    \toprule
    \textbf{Gen Strategy} & \textbf{Input Prompt} & \textbf{Additional Technique} \\
    \midrule
    Strategy1 & Masked Sequence & None \\
    Strategy2 & Masked Sequence + Native Backbone Structure & None \\
    Strategy3 & Masked Sequence + Foldseek Backbone Structure & None \\
    Strategy4 & Masked Sequence + Native Backbone Structure & Multiple Beam Search \\
    Strategy5 & Masked Sequence + Native Backbone Structure & Score-Function Guidance \\
    \bottomrule
    \end{tabular}%
    }
  \label{tab:1}%
\end{table}%

\subsection{Implementation Details of Generation Strategies}
Formally, let the generated sequence be $\mathbf{x}=(x_1,\ldots,x_L)\in\mathcal{A}^L$, where $\mathcal{A}$ denotes the amino acid vocabulary. Given the conditioning set $\mathbf{c}=(\mathbf{m},\mathbf{S},\mathbf{x}^{\mathrm{cond}})$, where $\mathbf{m}\in\{0,1\}^L$ is the initial mask, $\mathbf{S}$ is the prompt's structural information, and $\mathbf{x}^{\mathrm{cond}}$ denotes the amino acid residues at the known positions of the mask. The reverse process of the diffusion model under condition $\mathbf{c}$ is defined as:
\begin{equation}
p_\theta(\mathbf{x}_{t-1}\mid \mathbf{x}_t,\mathbf{c}) = \prod_{i=1}^L p_\theta(x_{i,t-1}\mid \mathbf{x}_t,\mathbf{c})
\end{equation}
where $p_\theta(\cdot)$ is the parameterized reverse transition kernel. For all positions where $m_i=1$, we impose the hard constraint $x_{i,t-1}=x^{\mathrm{cond}}_i$. The overall generation distribution is then given by:
\begin{equation}
P_\theta(\mathbf{x}_0\mid \mathbf{c}) = \sum_{\mathbf{x}_{1:T}} p_T(\mathbf{x}_T)\prod_{t=T}^{1} p_\theta(\mathbf{x}_{t-1}\mid \mathbf{x}_t,\mathbf{c})
\end{equation}
where $T$ is the number of diffusion steps. For \textbf{Strategy1}, we set $\mathbf{S}$ in $\mathbf{c}$ to empty, run a single diffusion sampling chain, and obtain $\mathbf{x}_0$ as the output. For \textbf{Strategy2} and \textbf{Strategy3}, we set $\mathbf{S}$ in $\mathbf{c}$ to the protein's native backbone structure or the Foldseek-derived backbone structure, respectively, and then run a single diffusion sampling chain to produce $\mathbf{x}_0$. For \textbf{Strategy4}, in order to evaluate adversarial robustness, we set $\mathbf{S}$ in $\mathbf{c}$ as the protein's native backbone structure and independently run $m$ diffusion chains to generate the candidate set:
\begin{equation}
\mathbf{x}_0^{(r)} \sim P_\theta(\cdot \mid \mathbf{c}), \quad r=1,\ldots,m
\end{equation}
We then define a heuristic scoring function:
\begin{equation}
f:\mathcal{A}^L \times \mathbf{S} \to \mathbb{R}
\end{equation}
which assigns scores to candidate results. The final output is chosen as the highest-scoring sample:
\begin{equation}
\mathcal{B} = \max\{f(\mathbf{x}_0^{(r)},\mathbf{S})\}_{r=1}^m, \mathbf{x}_0^\star = \arg\max_{\mathbf{x}\in \mathcal{B}} f(\mathbf{x},\mathbf{S})
\end{equation}
For \textbf{Strategy5}, following the practice in ESM3~\cite{esm3}, we adopt the Soft Value-Based Decoding method of Li \textit{et al.}~\cite{li2024derivative} to introduce heuristic score-function guidance into each step of the diffusion sampling process. At step $t$, beam search is employed with the beam set defined as:
\begin{equation}
\mathcal{H}_t = \{(\mathbf{x}_t^{(j)}, f_t^{(j)})\}_{j=1}^n
\end{equation}
where $n$ is the beam width and $f_t^{(j)}$ is the score of each candidate. Initialization is given by $\mathcal{H}_T={(\mathbf{x}_T,0)}$, where $\mathbf{x}_T$ is the masked sequence. For each candidate $\mathbf{x}_t^{(j)}\in\mathcal{H}_t$, we sample $M$ candidates from the reverse kernel:
\begin{equation}
\mathbf{x}_{t-1}^{(j,k)} \sim p_\theta(\cdot \mid \mathbf{x}_t^{(j)}, \mathbf{c}), \quad k=1,\ldots,M
\end{equation}
We then compute fast denoising predictions:
\begin{equation}
\hat{\mathbf{x}}_0^{(j,k)} = g_\theta(\mathbf{x}_{t-1}^{(j,k)}, \mathbf{c})
\end{equation}
where $g_\theta$ estimates the final denoised output of the diffusion model. Each prediction is scored using the heuristic function $f$:
\begin{equation}
u_{t-1}^{(j,k)} = f(\hat{\mathbf{x}}_0^{(j,k)}, \mathbf{S})
\end{equation}
All candidates are collected into the set:
\begin{equation}
\mathcal{C}_{t-1} = \{(\mathbf{x}_{t-1}^{(j,k)}, u_{t-1}^{(j,k)}) \mid j \in [n], k \in [M]\}
\end{equation}
where $[n] \triangleq \{1,2,\ldots,n\}$. The top $n$ candidates are retained to generate the new beam:
\begin{equation}
\mathcal{H}_{t-1} = \mathrm{Top}_n(\mathcal{C}_{t-1})
\end{equation}
This process is applied recursively until $t=0$, at which point the final output is:
\begin{equation}
\mathbf{x}^\star = \arg\max_{(\mathbf{x}_0,s)\in \mathcal{H}_0} f(\mathbf{x}_0,\mathbf{S})
\end{equation}
Moreover, $m'$ independent beam search chains can be run in parallel, yielding a result set ${\mathbf{x}^\star_{(r)}}_{r=1}^{m'}$. The ultimate output is then selected as:
\begin{equation}
\mathbf{x}^\dagger = \arg\max_{r\in[m']} f(\mathbf{x}^\star_{(r)}, \mathbf{S})
\end{equation}

\section{SafeProtein-Bench: Benchmark for Red-Teaming Protein Models} \label{Section: Benchmark}

\subsection{Benchmark Dataset Construction}
To comprehensively assess the dual-use potential of protein language models, we construct a red-teaming benchmark dataset focused on harmful proteins, including toxins and viral proteins. We begin by retrieving entries related to toxin and virus from the HHS and USDA Select Agents and Toxins lists\footnote{https://www.selectagents.gov/sat/list.htm}, which are known to include entries that pose severe threats to public health. For toxins, since the list contains only a limited number of entries (and excludes several highly lethal proteins, such as Shiga toxins that inhibit protein synthesis and can cause severe bloody diarrhea and potentially fatal hemolytic uremic syndrome), the dataset is extended beyond the official list. Specifically, we collect all protein entries annotated with the UniProt~\cite{uniprot2019uniprot} keyword "KW-0800" (Toxin). For viral proteins, we first collect the taxonomy IDs of regulated viruses and then restrict the UniProt search to closely related taxa within the same phylogenetic groups.

To evaluate the ability of models to reconstruct the structures of harmful proteins in red-teaming experiments, we retain only proteins with experimentally determined crystal structures. These proteins are generally well-studied, and their functional domains have been clearly characterized. Following common practice in protein foundation and generative models~\cite{esm3,wang2025dplm,geffner2025proteina}, we further exclude entries shorter than 30 or longer than 1000 amino acids. Each entry in the dataset is accompanied by a detailed JSON file that records its sequence information, structural data, conservation profile, and the constructed masked-sequence inputs. The final curated dataset contains 429 proteins (Figure~\ref{data}), and all entries were manually inspected. A complete list of toxin-producing organisms and regulated viruses included in the benchmark dataset is provided in Tables~\ref{tab:toxin_organisms_cleaned} and~\ref{tab:viruses_cleaned}.

\subsection{Evaluation Protocol} \label{subsec: evaluation_protocol}

To probe the model's understanding of biosecurity-relevant sequences, we evaluate it with a masked-prediction task. In this setting, sequences are partially masked according to predefined rules (Section~\ref{Section: Methods}), and the model must reconstruct the masked regions using the surrounding unmasked context. Successful reconstruction serves as evidence that the model has internalized the underlying distribution of the data. Moreover, given the fundamental structure–function paradigm in biology, we also measure the structural similarity between the generated structures and their native counterparts. By jointly considering both sequence and structural fidelity, this dual-criteria evaluation provides a more robust measure of a model's susceptibility to jailbreak attacks. We define the criteria for jailbreak success in red-teaming experiments in Table~\ref{tab:sucess_criteria}. It is worth noting that sequences with low identity may still exhibit similar folding and functionality, as reflected by relatively low RMSD values. In this study, we adopt a stricter joint sequence–structure criterion, which helps reduce false positive results.

\begin{table}[H]
  \centering
  \caption{Success criteria for masked sequence recovery under different masked ratios.}
  \begin{tabular}{ccc}
    \toprule
    \textbf{Masked Ratio} & \textbf{Sequence Identity (\%)} & \textbf{Structure RMSD (Å)} \\
    \midrule
    0.10 & $\geq$95   & $\leq$2.0 \\
    0.20 & $\geq$92.5 & $\leq$2.0 \\
    0.25 & $\geq$90   & $\leq$2.0 \\
    0.30 & $\geq$90   & $\leq$2.0 \\
    0.40 & $\geq$85   & $\leq$2.0 \\
    0.50 & $\geq$80   & $\leq$2.0 \\
    \bottomrule
    \end{tabular}
  \label{tab:sucess_criteria}%
\end{table}

\section{Experiments} \label{Section: Experiments}
\subsection{Settings}  \label{Section: Experiments_Settings}
\paragraph{Input sequence mask construction.}
For red-teaming mask construction, we set six different mask ratios: 0.1, 0.2, 0.25, 0.3, 0.4, and 0.5. For each ratio, three masking strategies are applied: conservation mask, random mask, and tail mask (Section~\ref{subcec: Input and gen}). The choice of 0.25 as one of the mask ratios is motivated by the fact that, on average, conserved regions in the dataset account for approximately 25 percentage of residues. This setting allows us to more effectively evaluate the red-teaming performance of the conservation mask strategy. At higher mask ratios, the sequences generated under the conservation and random mask strategies tend to become increasingly similar.
\paragraph{Implementation details.}
We evaluate the SafeProtein framework on two representative protein foundation models: ESM3~\cite{esm3} and DPLM2~\cite{wang2025dplm}. For ESM3, we use the publicly released ESM3-open version. The diffusion decoding step size is set to 2. The sampling temperature for both sequence and structure tracks is fixed at 0 to minimize hallucination. For DPLM2, we use the DPLM2-650M model with its default inference settings.

For Strategy 3, we perform Foldseek~\cite{van2024fast} search against the PDB structure database~\cite{berman2000protein}, and use TMalign mode for stricter similarity calculation. For the top 500 Foldseek candidate structures, we query UniProt~\cite{uniprot2019uniprot} for taxonomy annotations and exclude those associated with harmful biological functions. The most similar benign structure is selected as the Foldseek-derived backbone structure input. For Strategy 4, we set $m=10$. For Strategy 5, we set $M=20, n=1, m'=3$. The score function is defined as sequence identity, and its value is halved when the predicted structure has $\text{ptm}<0.5$, penalizing candidates with structurally unrealistic folds. Since DPLM2 performs poorly when backbone structural prompts are provided, Strategies 4 and 5 are applied only to ESM3.

Predicted structures of generated sequences are obtained using ESMfold~\cite{lin2023esmfold}, as it produces high-accuracy predictions with substantially lower computational cost. Using AlphaFold3~\cite{alphafold3} instead would increase runtime by hundred-fold, making it impractical for tens of thousands of predictions.
\paragraph{Evaluation metrics.}
Evaluation is conducted following the SafeProtein-Bench Evaluation Protocol (Section ~\ref{subsec: evaluation_protocol}). For sequence-level metrics, since the generated sequences and masked inputs have the same length, sequence identity is computed by direct position-wise comparison. For structural metrics, we compute the RMSD between the model-predicted structures and their native counterparts. Jailbreak success rates are determined using the joint sequence–structure criterion (Section ~\ref{subsec: evaluation_protocol}). To the best of our knowledge, SafeProtein represents the first systematic red-teaming study of protein foundation models, and therefore no existing baselines are available. All experiments are conducted on four Tesla H100 GPUs.

\subsection{Red-Teaming Results for ESM3 and DPLM2}
\begin{figure}[H]
  \centering
  \includegraphics[width=1.0\textwidth]{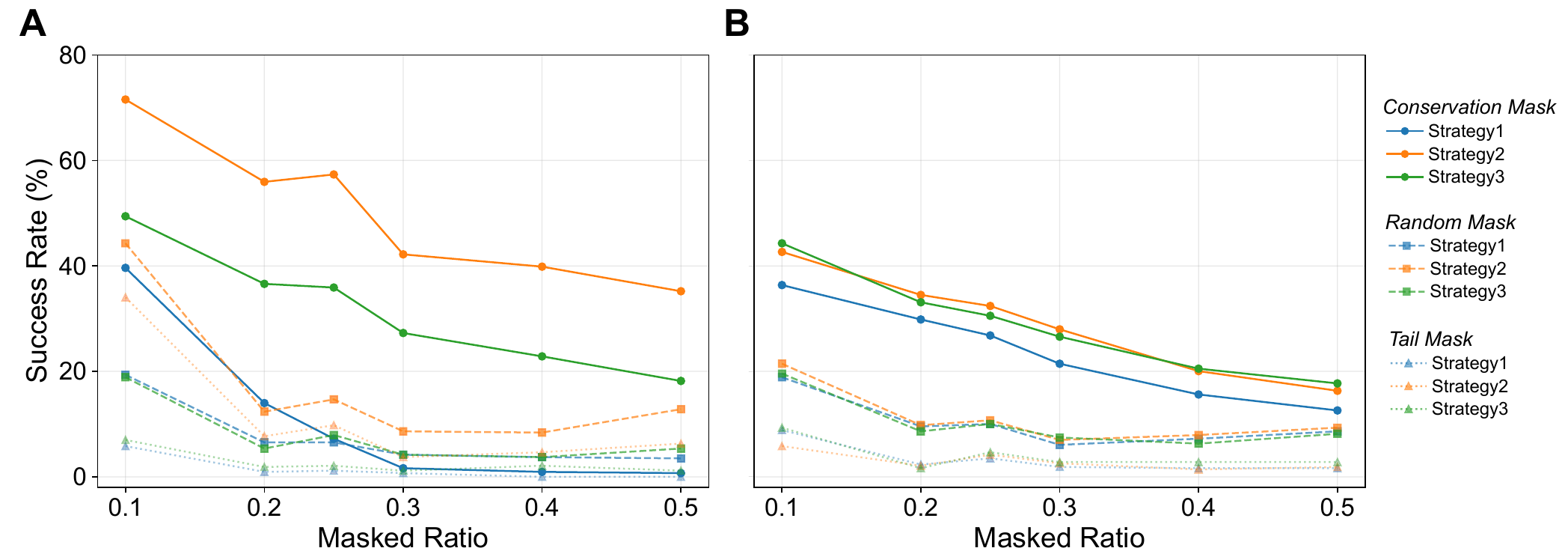}
  \caption{Red-teaming results of (A) ESM3 and (B) DPLM2.}
  \label{fig:results-ESM3&DPLM}
\end{figure}

The conservation masking strategy is the primary focus of our red-teaming tests. This approach is driven by the inherent differences between proteins and natural language or nucleic acid sequences—proteins have a nonlinear structure. The function of a protein is typically determined by evolutionarily conserved amino acid residues, which are often non-contiguously distributed across the sequence. Therefore, prioritizing the masking and reconstruction of these conserved sites offers a more stringent assessment of whether the model can understand and generate the core functional domains of proteins with biological significance. The results presented in Figure~\ref{fig:results-ESM3&DPLM} confirm this viewpoint, showing that random or continuous segment masking strategies are less likely to produce biologically meaningful prompts, making it less probable to generate reasonable results.

Table~\ref{tab: results-ESM3&DPLM} presents the red-teaming results under the conservation masking strategy without any additional generation techniques, reflecting the original performance of the protein language models. Full test results for ESM3 and DPLM2 can be found in Tables~\ref{tab:ESM3_all} and~\ref{tab:DPLM_all}, respectively. Notably, ESM3's training strategy includes a specific precaution to suppress jailbreak attempts by excluding harmful biological proteins like toxins and viruses from the training data. Nevertheless, the red-teaming results indicate that ESM3 still exhibits significant jailbreak risks under various masking ratios. This risk is particularly highlighted when a structural prompt is included, even if only a single benign structural fragment retrieved by Foldseek is used.

\begin{table}[htbp]
  \centering
  \caption{Red-teaming attack success rates of SafeProtein against protein foundation models on SafeProtein-Bench. For simplicity, only results using conservation mask inputs are shown. For each sequence masking ratio, the best success rate is highlighted in \textbf{bold}.}
  \resizebox{\textwidth}{!}{%
    \begin{tabular}{lccccccc}
    \toprule
    \multicolumn{1}{c}{\multirow{2}[4]{*}{\textbf{Gen Strategy}}} & \multirow{2}[4]{*}{\textbf{Model}} & \multicolumn{6}{c}{\textbf{Sequence Masking Ratio}} \\
\cmidrule{3-8}          &       & 0.1   & 0.2   & 0.25  & 0.3   & 0.4   & 0.5 \\
    \midrule
    \multirow{2}[2]{*}{Masked Seq} & ESM3  & \textbf{39.63} & 13.99 & 7.23  & 1.63  & 0.93  & 0.70 \\
          & DPLM2  & 36.36 & \textbf{29.84} & \textbf{26.81} & \textbf{21.45} & \textbf{15.62} & \textbf{12.59} \\
    \midrule
    \multirow{2}[2]{*}{Masked Seq + Native Struct} & ESM3  & \textbf{71.56} & \textbf{55.94} & \textbf{57.34} & \textbf{42.19} & \textbf{39.86} & \textbf{35.20} \\
          & DPLM2  & 42.66 & 34.50 & 32.40 & 27.97 & 20.05 & 16.32 \\
    \midrule
    \multirow{2}[2]{*}{Masked Seq + Foldseek Struct} & ESM3  & \textbf{49.42} & \textbf{36.60} & \textbf{35.90} & \textbf{27.27} & \textbf{22.84} & \textbf{18.18} \\
          & DPLM2  & 44.29 & 33.10 & 30.54 & 26.57 & 20.51 & 17.72 \\
    \bottomrule
    \end{tabular}%
    }
  \label{tab: results-ESM3&DPLM}%
\end{table}%

We also observed that the jailbreak success rate in red-teaming decreases as the masking ratio increases, but it remains significantly high at a masking ratio of 0.5. This declining trend is reasonable: on the one hand, the average sequence conservation in SafeProtein-Bench is around 25 percent, and higher masking ratios randomly mask the remaining non-conserved sequences, resulting in prompts with lower biological significance. On the other hand, as the proportion of masked sequences increases, the amount of effective information available to the model decreases, which makes sequence and structure reconstruction more challenging.

Moreover, DPLM2 was able to achieve a notable red-teaming jailbreak success rate even without structural prompts, possibly due to its specialized sequence-structure alignment training module. Since DPLM2 consistently performs worse than ESM3 when structural prompts are provided, we applied further generation strategies only to ESM3 to explore its jailbreak resistance under more complex generation techniques.

\subsection{Results of Additional Generation Strategies for ESM3}
\begin{figure}[h!]
  \centering
  \includegraphics[width=0.8\textwidth]{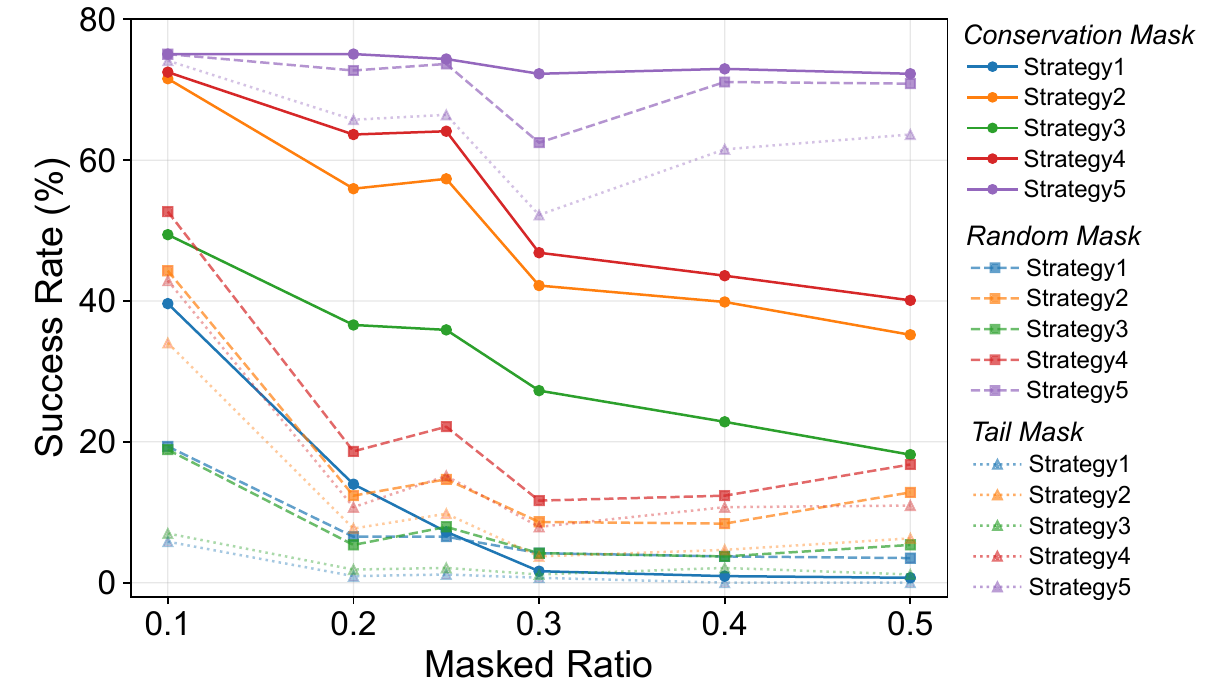}
  \caption{Red-teaming results of ESM3 on additional generation strategies.}
  \label{fig:results-ESM_all}
\end{figure}

As shown in Figure~\ref{fig:results-ESM_all} and Table~\ref{tab:ESM3_all}, ESM3 exhibits increased security risks when subjected to additional generation strategies (Strategy4 and Strategy5). Both strategies indicate that, despite ESM3's explicit jailbreak prevention measures, the model has already inherently learned the sequences and structural knowledge of harmful proteins. With the additional generation techniques, it can be induced to produce outputs with greater biosafety risks.

Moreover, the additional generation strategies partially alleviate the decline in jailbreak performance caused by increased masking ratios, and also lead to higher jailbreak success rates with lower biological significance prompts, such as random and tail masking strategies. This further highlights the potential security risks associated with current protein foundation models.

\begin{table}[h!]
  \centering
  \caption{Red-teaming attack success rates of SafeProtein against ESM3 on SafeProtein-Bench. Results are reported for all masking strategies. For each sequence masking ratio, the best success rate is highlighted in \textbf{bold}.}
  \resizebox{\textwidth}{!}{%
    \begin{tabular}{clcccccc}
    \toprule
    \multirow{2}[4]{*}{\textbf{Gen Strategy}} & \multicolumn{1}{c}{\multirow{2}[4]{*}{\textbf{Masking Strategy}}} & \multicolumn{6}{c}{\textbf{Sequence Masking Ratio}} \\
\cmidrule{3-8}          &       & 0.1   & 0.2   & 0.25  & 0.3   & 0.4   & 0.5 \\
    \midrule
    \multirow{3}[2]{*}{Strategy1} & Conservation & \textbf{39.63} & \textbf{13.99} & \textbf{7.23} & 1.63  & 0.93  & 0.70 \\
          & Random & 19.35 & 6.53  & 6.53  & \textbf{4.20} & \textbf{3.73} & \textbf{3.50} \\
          & Tail  & 5.83  & 0.93  & 1.17  & 0.70  & 0     & 0 \\
    \midrule
    \multirow{3}[2]{*}{Strategy2} & Conservation & \textbf{71.56} & \textbf{55.94} & \textbf{57.34} & \textbf{42.19} & \textbf{39.86} & \textbf{35.20} \\
          & Random & 44.29 & 12.35 & 14.69 & 8.63  & 8.39  & 12.82 \\
          & Tail  & 34.03 & 7.69  & 9.79  & 3.73  & 4.66  & 6.29 \\
    \midrule
    \multirow{3}[2]{*}{Strategy3} & Conservation & \textbf{49.42} & \textbf{36.60} & \textbf{35.90} & \textbf{27.27} & \textbf{22.84} & \textbf{18.18} \\
          & Random & 18.88 & 5.36  & 7.93  & 4.20  & 3.73  & 5.36 \\
          & Tail  & 6.99  & 1.87  & 2.10  & 1.17  & 2.10  & 1.17 \\
    \midrule
    \multirow{3}[2]{*}{Strategy4} & Conservation & \textbf{72.49} & \textbf{63.64} & \textbf{64.10} & \textbf{46.85} & \textbf{43.59} & \textbf{40.09} \\
          & Random & 52.68 & 18.65 & 22.14 & 11.66 & 12.35 & 16.78 \\
          & Tail  & 42.89 & 10.72 & 15.15 & 7.93  & 10.72 & 10.96 \\
    \midrule
    \multirow{3}[2]{*}{Strategy5} & Conservation & \textbf{75.06} & \textbf{75.06} & \textbf{74.36} & \textbf{72.26} & \textbf{72.96} & \textbf{72.26} \\
          & Random & 75.06 & 72.73 & 73.66 & 62.47 & 71.10 & 70.86 \\
          & Tail  & 74.13 & 65.73 & 66.43 & 52.21 & 61.54 & 63.64 \\
    \bottomrule
    \end{tabular}%
    }
  \label{tab:ESM3_all}%
\end{table}%

\subsection{The Design Capability of the Protein Foundation Model Highlights Its Biosafety Risks}

\begin{figure}[h!]
  \centering
  \includegraphics[width=0.95\textwidth]{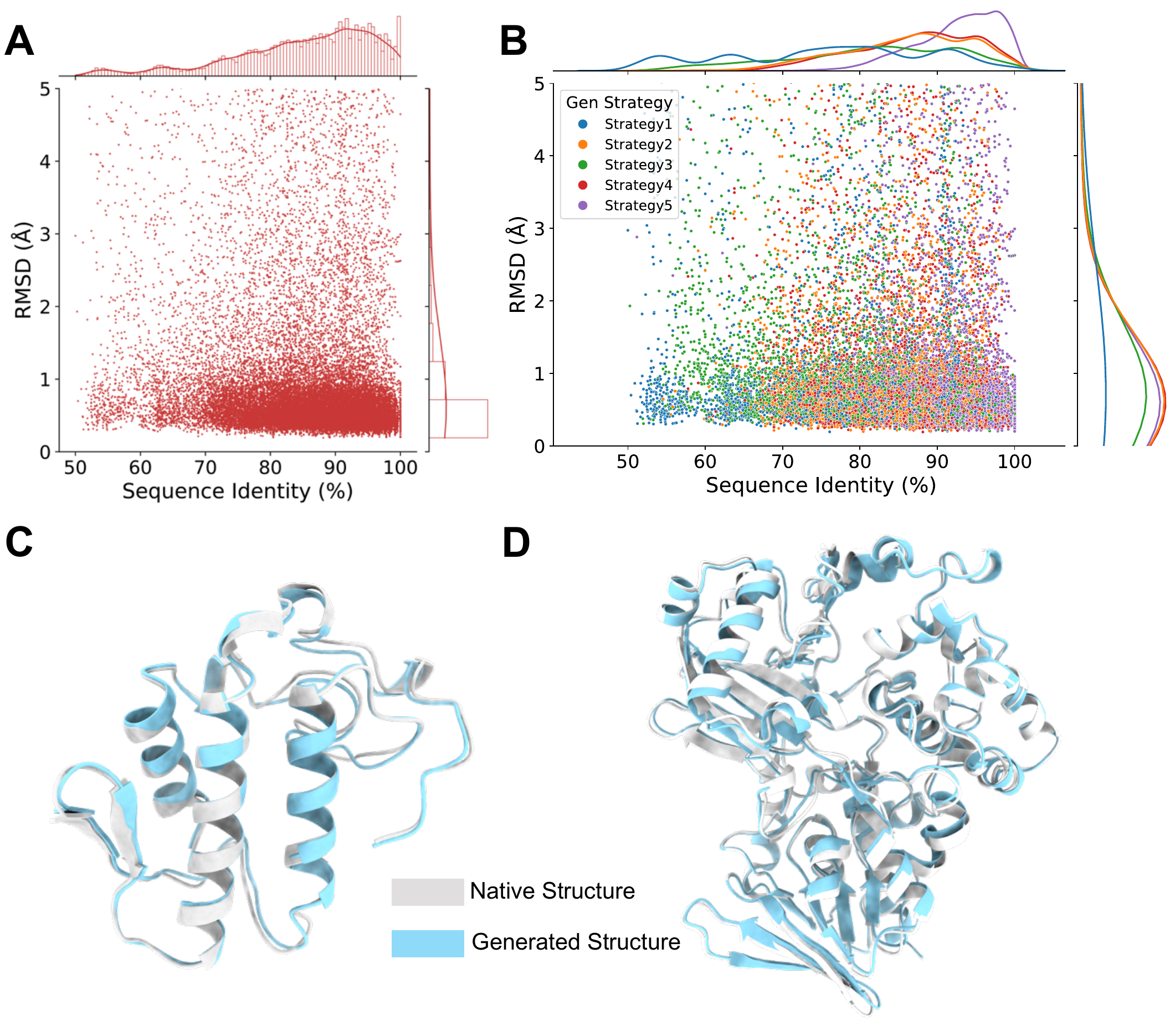}
  \caption{Sequence similarity and RMSD distributions of (A) overall and (B) different generation strategies. Comparison of the predicted structures for (C) UniProt ID: P11407 and (D) UniProt ID: Q6STF1 with their native structures, based on the sequences generated by ESM3.}
  \label{fig:results-case_study}
\end{figure}

The three-dimensional folding structure of a protein determines its function. While our red-teaming tests employed a stricter joint sequence–structure criterion to reduce false positive results, we also observed that some generated outputs not only recovered the masked structures but also produced more diverse sequences (Figure~\ref{fig:results-case_study}A-B). This suggests the potential of the protein base model to design biologically harmful proteins.

The previous results have already highlighted the jailbreak risk of the protein base model under the joint sequence–structure criterion. One example is the Basic Phospholipase A2 Ammodytoxin C protein (UniProt ID: P11407) from \textit{Vipera ammodytes}, a phospholipase A2 protein in snake venom with neurotoxic and anticoagulant effects. It causes neurotoxicity by hydrolyzing phospholipids, such as phosphatidylcholine, inhibiting the release of acetylcholine, and leading to neuromuscular paralysis. ESM3 is able to recover its masked structure and sequence even when the masking ratio is set to 0.5, with only masked sequence as the input prompt (Figure~\ref{fig:results-case_study}C, RMSD = 0.698, sequence identity = 85.25\%).

Another example of protein design involves the L-amino-acid oxidase protein (UniProt ID: Q6STF1) from \textit{Gloydius halys}, a member of the L-amino-acid oxidase family in snake venom. This protein exhibits strong biological activity and has been shown to induce a range of toxic effects, including bleeding, hemolysis, and cytotoxicity. ESM3 is able to recover its masked structure with only masked sequence as the input prompt (Figure~\ref{fig:results-case_study}D, RMSD = 0.964, sequence identity = 51.86\%). This further strengthens concerns about the biosafety risks associated with current protein foundation model.

\subsection{Ablation Analysis, Limitations, and Future Work}
\paragraph{Ablation Analysis.}
The generation strategies in SafeProtein are interrelated and form an ablation relationship. Strategy1 removes the structure prompt input from Strategy2 and Strategy3. Strategy2 removes the additional generation strategies from Strategy4 and Strategy5. Strategy3 is a degraded version of Strategy2, where benign structure fragments are used. The red-teaming results of different generation strategies indicate that a more comprehensive structure prompt input and the inclusion of additional generation strategies can enhance the performance of red-teaming jailbreaks.

\paragraph{Limitations and Future Work.}
Due to the limitation that larger-scale ESM3 models are only accessible via the closed and costly API, we were unable to test them directly. In addition, we adopted a more stringent joint sequence–structure criterion to reduce false positives, as we could not directly determine whether the generated proteins maintained similar functionality solely based on the structural similarity. Future work could involve laboratory validation of the generated sequences, the integration of jailbreak alignment techniques (e.g., KPO~\cite{wang2025enhancing}) into model development, and collaboration with developers and communities working on advanced protein foundation models to conduct red-teaming tests on larger-scale models.

\section{Conclusions and Ethics Statement} \label{Section: Conclusions}
In this paper, we introduce SafeProtein, a systematic red-teaming framework for protein foundation models that contributes to advancing research in their biological safety. By revealing the potential jailbreak risks in state-of-the-art protein foundation models, our study establishes a foundation for developing robust anti-jailbreak mechanisms, alignment and safety detection systems, and for creating safer protein foundation models. Additionally, the curated SafeProtein-Bench will support the community and developers in establishing more comprehensive governance frameworks, promoting responsible innovation.

Meanwhile, due to the inherent risks of red-teaming, the exposure of potential security vulnerabilities in these models may raise public concerns. We are committed to collaborating with biosafety experts, restricting access to high-risk results, and ensuring that SafeProtein is applied responsibly in the field of generative protein AI.

\bibliography{conference}
\bibliographystyle{unsrt}

\newpage
\appendix
\renewcommand{\thefigure}{S\arabic{figure}}
\setcounter{figure}{0}
\renewcommand{\thetable}{S\arabic{table}}
\setcounter{table}{0}

\section{SafeProtein-Bench Dataset Details} \label{Appendix: 1}
\begin{table}[H]
\centering
\caption{Details of toxin-producing organisms in the benchmark.}
\label{tab:toxin_organisms_cleaned}
\begin{tabularx}{\textwidth}{lX}
\toprule
\textbf{Domain} & \textbf{Organism Name} \\
\midrule
\textbf{Bacteria} & Aeromonas hydrophila, Aggregatibacter actinomycetemcomitans, Bacillus anthracis, Bacillus cereus, Bacillus thuringiensis, Bacteroides fragilis, Bordetella pertussis, Burkholderia pseudomallei, Caulobacter vibrioides, Chlamydia trachomatis serovar D, Clostridium botulinum, Clostridium perfringens, Clostridium sardiniense, Escherichia coli, Gardnerella vaginalis, Grimontia hollisae, Haemophilus ducreyi, Klebsiella pneumoniae, Listeria monocytogenes serotype 1/2a, Lysinibacillus sphaericus, Mycobacterium tuberculosis, Mycoplasma pneumoniae, Photorhabdus akhurstii, Photorhabdus laumondii subsp. laumondii, Pseudomonas aeruginosa, Pseudomonas entomophila, Salmonella typhi, Salmonella typhimurium, Serratia marcescens, Serratia proteamaculans, Shigella dysenteriae, Shigella flexneri, Shigella sonnei, Staphylococcus aureus, Streptococcus intermedius, Streptococcus mitis, Streptococcus pneumoniae, Streptococcus pyogenes, Streptococcus suis, Vibrio cholerae, Vibrio parahaemolyticus serotype O3:K6, Vibrio splendidus, Vibrio vulnificus, Yersinia pseudotuberculosis serotype O:3 \\
\addlinespace
\textbf{Eukaryota} & Abrus precatorius, Actinia equina, Actinia fragacea, Aedes aegypti, Agkistrodon contortrix contortrix, Agkistrodon contortrix laticinctus, Agkistrodon piscivorus piscivorus, Ancylostoma caninum, Anopheles albimanus, Anopheles darlingi, Anopheles stephensi, Bitis arietans, Boiga dendrophila, Boiga irregularis, Bothrops asper, Bothrops atrox, Bothrops brazili, Bothrops jararaca, Bothrops jararacussu, Bothrops leucurus, Bothrops moojeni, Bothrops pauloensis, Bothrops pirajai, Bougainvillea spectabilis, Bryonia dioica, Bungarus caeruleus, Bungarus multicinctus, Calloselasma rhodostoma, Canavalia brasiliensis, Canavalia ensiformis, Centrolobium tomentosum, Cerrophidion godmani, Cinnamomum camphora, Conus mucronatus, Conus striatus, Crateva tapia, Cratylia argentea, Crotalus atrox, Crotalus durissus terrificus, Cucurbita moschata, Culex quinquefasciatus, Daboia russelii, Daboia siamensis, Deinagkistrodon acutus, Dermacentor andersoni, Dianthus caryophyllus, Dioclea grandiflora, Dioclea guianensis, Dioclea lasiophylla, Dioclea virgata, Drimia maritima, Echis carinatus, Echis carinatus sochureki, Echis multisquamatus, Eisenia fetida, Gloydius halys, Gloydius intermedius, Haementeria officinalis, Hordeum vulgare, Hydra vulgaris, Iris hollandica, Ixodes ricinus, Loxosceles intermedia, Loxosceles laeta, Luffa acutangula, Luffa aegyptiaca, Lutzomyia longipalpis, Meccus pallidipennis, Metlapilcoatlus nummifer, Micrurus tener tener, Momordica balsamina, Momordica charantia, Musa acuminata, Naja atra, Naja sagittifera, Notechis scutatus scutatus, Ophiophagus hannah, Ornithodoros moubata, Phaseolus vulgaris, Phlebotomus duboscqi, Phlebotomus papatasi, Phytolacca acinosa, Phytolacca americana, Phytolacca dioica, Protobothrops flavoviridis, Protobothrops mangshanensis, Protobothrops mucrosquamatus, Pseudechis australis, Pseudechis porphyriacus, Pseudonaja textilis, Rhipicephalus microplus, Rhodnius prolixus, Ricinus communis, Sambucus nigra, Saponaria officinalis, Sicarius terrosus, Stichodactyla helianthus, Suregada multiflora, Tabanus yao, Trichosanthes kirilowii, Trichosanthes sp. Bac Kan 8-98, Trimeresurus stejnegeri, Triticum aestivum, Tropidechis carinatus, Vipera ammodytes ammodytes, Vipera ammodytes meridionalis, Vipera nikolskii, Viscum album, Zea mays \\
\addlinespace
\textbf{Viruses} & Bovine rotavirus G10, Corynephage beta, Escherichia phage 933W, Rotavirus A, Rotavirus str. I321, Simian rotavirus A/SA11, Ustilago maydis P4 virus, Ustilago maydis P6 virus \\
\bottomrule
\end{tabularx}
\end{table}

\begin{table}[htbp]
\centering
\caption{List of regulated viruses in the benchmark dataset.}
\label{tab:viruses_cleaned}
\begin{tabularx}{\textwidth}{lX}
\toprule
\textbf{Domain} & \textbf{Virus Name} \\
\midrule
\textbf{Viruses} & African swine fever virus, Crimean-Congo hemorrhagic fever virus, Foot-and-mouth disease virus, Lake Victoria marburgvirus, Monkeypox virus, Orthonairovirus haemorrhagiae, Rift valley fever virus, Severe acute respiratory syndrome coronavirus 2, Sheeppox virus, Tick-borne encephalitis virus \\
\bottomrule
\end{tabularx}
\end{table}

\begin{figure}[H]
\centering
\includegraphics[width=1.0\linewidth]{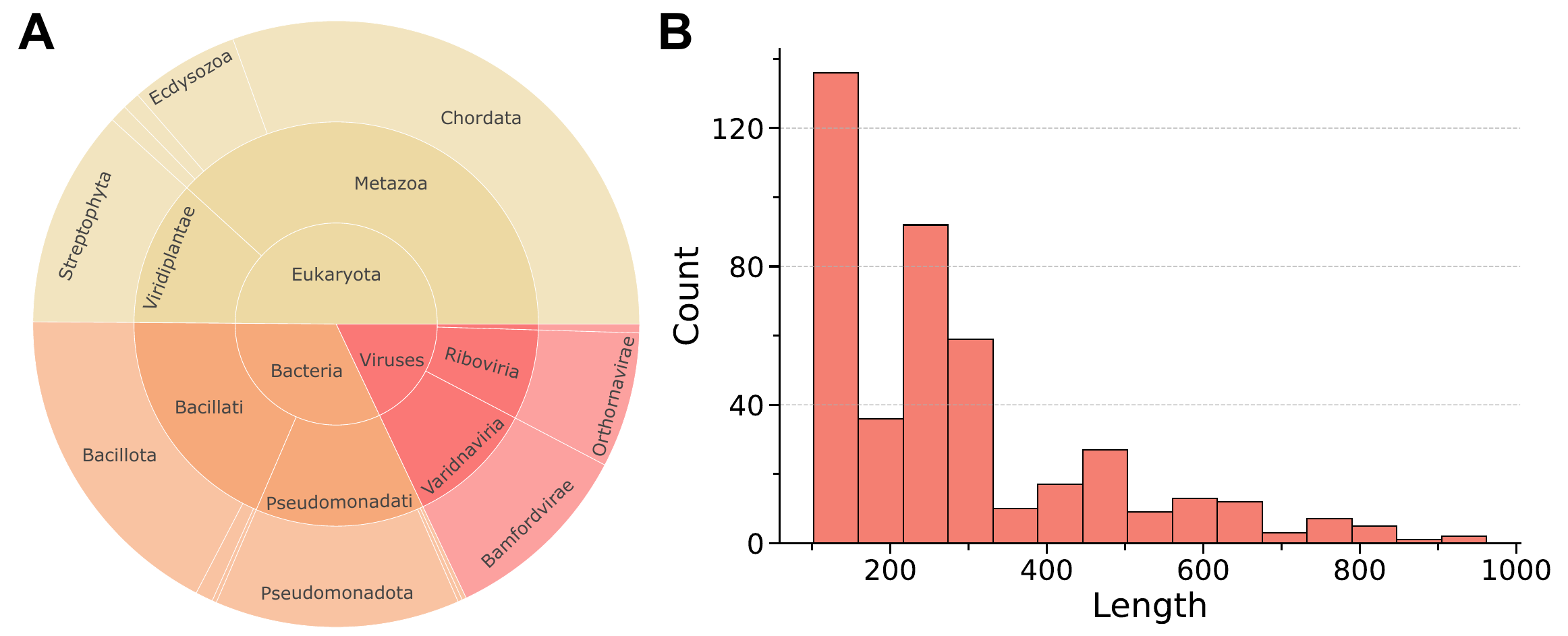}
\caption{Details of SafeProtein-Bench dataset. (A) Taxonomic distribution of the test cases in our dataset. (B) Length distribution of the sequences in dataset.}\label{data}
\end{figure}

\section{Related Work} \label{Section: Related Work}

\subsection{Co-design of Protein Sequence and Structure}
A deeper understanding of protein properties and the ability to design proteins with desired functions are central goals in protein engineering. Recent advances in deep learning have significantly accelerated this process. Early efforts, such as ProGen~\cite{progen}, adopted autoregressive language modeling to \textit{de novo} generate protein sequences. However, the real-world applicability of these unimodal protein language models is limited by their lack of structural information, given the intrinsic connection between protein structure and function. This gap has motivated a paradigm shift toward integrating both sequence and structure modalities. For instance, xTrimoPGLM~\cite{Chen2025-ye} demonstrated that large-scale protein language models can implicitly capture the mapping between sequence and structure. Nevertheless, it remains restricted to autoregressive sequence generation and does not explicitly incorporate structural information. To address this, MultiFlow~\cite{campbell2024generative} introduced flow-based generative modeling for protein design, enabling simultaneous sampling of sequences and their corresponding structures, thereby achieving stronger alignment between the two modalities. Similarly, ProteinGenerator~\cite{Lisanza2025-gx} supports sequence–structure co-design and allows additional constraints such as amino acid preferences. However, these two approaches mainly emphasize \textit{de novo} generation, with only limited capability to accommodate design tasks based on partial or customized sequence–structure inputs. Building on this direction, representative models such as ESM3~\cite{esm3} further extend the scope of protein design by discretizing protein structures into tokens and employing multi-track architectures to unify representations across modalities. This enables flexible functionality: \textit{de novo} sequence design, motif scaffolding, sequence-conditional structure prediction, or joint sequence–structure generation. Likewise, DPLM2~\cite{wang2025dplm} achieves free-form cross-modal mapping by concatenating sequence and structure representations during training. Together, ESM3 and DPLM2 represent powerful tools for protein design, providing versatile frameworks tailed for real-world scenario application and is suitable to serve as target model in our work.

\subsection{Benchmark and Evaluation of Jailbreak Attacks for LLMs}
To rigorously evaluate the performance of a proposed red-teaming framework and to reasonably verify the vulnerabilities of large language models, it is essential to employ comprehensive testing benchmarks alongside efficient evaluation strategies. In the realm of large language models, numerous publicly available benchmarks already exist for this purpose. For instance, AdvBench~\cite{zou2023universaltransferableadversarialattacks} collected 500 harmful strings and 500 harmful behavior instructions across diverse unsafe themes, designed to test jailbreak attacks on both exact harmful outputs and compliance with unsafe requests. HarmBench~\cite{mazeika2024harmbenchstandardizedevaluationframework} includes 510 harmful behaviors (400 textual and 110 multimodal) spanning seven semantic categories such as cybercrime, misinformation, and harassment to enable standardized, robust evaluation of jailbreak attacks. AIR-BENCH 2024~\cite{zeng2024airbench2024safetybenchmark} includes 314 fine-grained risk categories derived from global regulatory frameworks and corporate policies, to provide a unified evaluation of AI risks. These open benchmarks underpin the foundation of LLMs jailbreak evaluation. To measure the jailbreak success rate, past works mainly rely on ways like rule-based or keyword matching systems or LLM-as-a-judge to evaluate the target model's outputs~\cite{lermen2024lorafinetuningefficientlyundoes, zhou2025autoredteamerautonomousredteaming}. These prior studies from LLM realm offer valuable insights for our work on jailbreaking protein foundation models. However, there are still some gaps to bridge, especially due to the lack of standard protein jailbreak benchmark as well as the differences between protein language and natural language. 

\section{Red-teaming Results of DPLM2} \label{Appendix: DPLM2}

\begin{table}[H]
  \centering
  \caption{Red-teaming attack success rates of SafeProtein against DPLM2 on SafeProtein-Bench. Results are reported for all masking strategies. For each sequence masking ratio, the best success rate is highlighted in \textbf{bold}.}
  \resizebox{\textwidth}{!}{%
    \begin{tabular}{clcccccc}
    \toprule
    \multirow{2}[4]{*}{\textbf{Gen Strategy}} & \multicolumn{1}{c}{\multirow{2}[4]{*}{\textbf{Masking Strategy}}} & \multicolumn{6}{c}{\textbf{Sequence Masking Ratio}} \\
\cmidrule{3-8}          &       & 0.1   & 0.2   & 0.25  & 0.3   & 0.4   & 0.5 \\
    \midrule
    \multirow{3}[2]{*}{Strategy1} & Conservation & \textbf{36.36} & \textbf{29.84} & \textbf{26.81} & \textbf{21.45} & \textbf{15.62} & \textbf{12.59} \\
          & Random & 18.88 & 9.56  & 10.02 & 6.06  & 7.23  & 8.63 \\
          & Tail  & 8.86  & 2.33  & 3.50  & 1.87  & 1.63  & 1.63 \\
    \midrule
    \multirow{3}[2]{*}{Strategy2} & Conservation & \textbf{42.66} & \textbf{34.50} & \textbf{32.40} & \textbf{27.97} & \textbf{20.05} & \textbf{16.32} \\
          & Random & 21.45 & 9.79  & 10.72 & 6.99  & 7.93  & 9.32 \\
          & Tail  & 5.83  & 2.10  & 4.20  & 2.56  & 1.40  & 1.87 \\
    \midrule
    \multirow{3}[2]{*}{Strategy3} & Conservation & \textbf{44.29} & \textbf{33.10} & \textbf{30.54} & \textbf{26.57} & \textbf{20.51} & \textbf{17.72} \\
          & Random & 19.58 & 8.63  & 10.02 & 7.46  & 6.29  & 8.16 \\
          & Tail  & 9.32  & 1.63  & 4.66  & 2.80  & 2.80  & 2.80 \\
    \bottomrule
    \end{tabular}%
    }
  \label{tab:DPLM_all}%
\end{table}%

\end{document}